\documentclass[11pt,letterpaper]{article}
\usepackage{naaclhlt2016}
\usepackage{times}
\usepackage{latexsym}
\usepackage{graphicx}
\usepackage{multirow}
\usepackage{siunitx}
\usepackage{url}
\usepackage{array,graphicx}
\usepackage{booktabs}
\usepackage{pifont}
\usepackage{changepage}
\usepackage[font=footnotesize,labelfont=bf]{caption}
\expandafter\def\expandafter\normalsize\expandafter{%
    \normalsize
    \setlength\abovedisplayskip{8pt}
    \setlength\belowdisplayskip{5pt}
    \setlength\abovedisplayshortskip{8pt}
    \setlength\belowdisplayshortskip{5pt}
}

\newcommand*\rot{\rotatebox{90}}
\newcommand*\OK{\ding{51}}

\naaclfinalcopy % Uncomment this line for the final submission
\def\naaclpaperid{***} %  Enter the naacl Paper ID here

\title{Learning Distributed Representations of Sentences from Unlabelled Data}

\author{Felix Hill\\
	    Computer Laboratory\\
	    University of Cambridge\\
	    {\tt \small felix.hill@cl.cam.ac.uk}
	  \And
	Kyunghyun Cho\\
  	Courant Institute of \\ 
        Mathematical Sciences\\
  	\& Centre for Data Science\\
  	New York University\\
  {\tt \small kyunghyun.cho@nyu.edu}
	  \And
	Anna Korhonen\\
	    Department of Theoretical \\ 
             \& Applied Linguistics \\
	    University of Cambridge\\
	    {\tt \small alk23@cam.ac.uk}}

\date{}

\begin{document}
\sisetup{tight-spacing=true}
\maketitle

\begin{abstract}
Unsupervised methods for learning distributed representations of words are ubiquitous in today's NLP research, but far less is known about the best ways to learn distributed phrase or sentence representations from unlabelled data. This paper is a systematic comparison of models that learn such representations. We find that the optimal approach depends critically on the intended application. Deeper, more complex models are preferable for representations to be used in supervised systems, but shallow log-linear models work best for building representation spaces that can be decoded with simple spatial distance metrics. We also propose two new unsupervised representation-learning objectives designed to optimise the trade-off between training time, domain portability and performance. 
  
\end{abstract}

\section{Introduction}

Distributed representations - dense real-valued vectors that encode the semantics of linguistic units - are ubiquitous in today's NLP research. For single-words or word-like entities, there are established ways to acquire such representations from naturally occurring (unlabelled) training data based on comparatively task-agnostic objectives (such as predicting adjacent words). These methods are well understood empirically \cite{baroni2014don} and theoretically \cite{levy2014neural}. The best word representation spaces reflect consistently-observed aspects of human conceptual organisation~\cite{hill2014simlex}, and can be added as features to improve the performance of numerous language processing systems \cite{collobert2011natural}. 

By contrast, there is comparatively little consensus on the best ways to learn distributed representations of phrases or sentences.\footnote{See the contrasting conclusions in \cite{mitchell2008vector,clark2007combining,baroni2014frege,milajevs2014evaluating} among others.} With the advent of deeper language processing techniques, it is relatively common for models to represent phrases or sentences as continuous-valued vectors. Examples include machine translation~\cite{sutskever2014sequence}, image captioning \cite{mao2014deep} and dialogue systems~\cite{serban2015building}. While it has been observed informally that the internal sentence representations of such models can reflect semantic intuitions~\cite{cho2014learning}, it is not known which architectures or objectives yield the `best' or most useful representations. Resolving this question could ultimately have a significant impact on language processing systems. Indeed, it is phrases and sentences, rather than individual words, that encode the human-like general world knowledge (or `common sense')~\cite{norman1972memory} that is a critical missing part of most current language understanding systems.

We address this issue with a systematic comparison of cutting-edge methods for learning distributed representations of sentences. We constrain our comparison to methods that do not require labelled data gathered for the purpose of training models, since such methods are more cost-effective and applicable across languages and domains. We also propose two new phrase or sentence representation learning objectives - \emph{Sequential Denoising Autoencoders} (SDAEs) and \emph{FastSent}, a sentence-level log-linear bag-of-words model. We compare all methods on two types of task - \emph{supervised} and \emph{unsupervised evaluations} - reflecting different ways in which representations are ultimately to be used. In the former setting, a classifier or regression model is applied to representations and trained with task-specific labelled data, while in the latter, representation spaces are directly queried using cosine distance.    

We observe notable differences in approaches depending on the nature of the evaluation metric. In particular, deeper or more complex models (which require greater time and resources to train) generally perform best in the supervised setting, whereas shallow log-linear models work best on unsupervised benchmarks. Specifically, SkipThought Vectors~\cite{kiros2015skip} perform best on the majority of supervised evaluations, but SDAEs are the top performer on paraphrase identification. In contrast, on the (unsupervised) SICK sentence relatedness benchmark, FastSent, a simple, log-linear variant of the SkipThought objective, performs better than all other models. Interestingly, the method that exhibits strongest performance across both supervised and unsupervised benchmarks is a bag-of-words model trained to compose word embeddings using dictionary definitions~\cite{hill2015learning}. Taken together, these findings constitute valuable guidelines for the application of phrasal or sentential representation-learning to language understanding systems.

\section{Distributed Sentence Representations}

To constrain the analysis, we compare neural language models that compute sentence representations from unlabelled, naturally-ocurring data, as with the predominant methods for word representations.\footnote{This excludes innovative supervised sentence-level architectures including \cite{socher2011semi,kalchbrenner2014convolutional} and many others.} Likewise, we do not focus on `bottom up' models where phrase or sentence representations are built from fixed mathematical operations on word vectors (although we do consider a canonical case - see CBOW below); these were already compared by~\newcite{milajevs2014evaluating}. Most space is devoted to our novel approaches, and we refer the reader to the original papers for more details of existing models. 

\subsection{Existing Models Trained on Text}
{\bf SkipThought Vectors} For consecutive sentences \(S_{i-1},S_i,S_{i+1}\) in some document, the {\bf SkipThought} model \cite{kiros2015skip} is trained to predict target sentences \(S_{i-1}\) and \(S_{i+1}\) given source sentence \(S_i\). As with all~\emph{sequence-to-sequence} models, in training the source sentence is `encoded' by a Recurrent Neural Network (RNN) (with Gated Recurrent uUnits~\cite{cho2014learning}) and then `decoded' into the two target sentences in turn. Importantly, because RNNs employ a single set of update weights at each time-step, both the encoder and decoder are sensitive to the order of words in the source sentence. 

For each position in a target sentence \(S_t\), the decoder computes a softmax distribution over the model's vocabulary. The cost of a training example is the sum of the negative log-likelihood of each correct word in the target sentences \(S_{i-1}\) and \(S_{i+1}\). This cost is backpropagated to train the encoder (and decoder), which, when trained, can map sequences of words to a single vector.

\vspace{5pt}\noindent {\bf ParagraphVector} \newcite{le2014distributed} proposed two log-linear models of sentence representation. The {\bf DBOW} model learns a vector \(\mathbf{s}\) for every sentence \(S\) in the training corpus which, together with word embeddings \(v_w\), define a softmax distribution optimised to predict words \(w \in S\) given \(S\). The \(v_w\) are shared across all sentences in the corpus. In the {\bf DM} model, \(k\)-grams of consecutive words \(\{w_i \dots w_{i+k} \in S\}\) are selected and \(\mathbf{s}\) is combined with \(\{v_{w_i} \dots v_{w_{i+k}} \}\) to make a softmax prediction (parameterised by additional weights) of \(w_{i+k+1}\). 

We used the Gensim implementation,\footnote{\scriptsize \url{https://radimrehurek.com/gensim/}} treating each sentence in the training data as a `paragraph' as suggested by the authors. During training, both DM and DBOW models store representations for every sentence (as well as word) in the training corpus. Even on large servers it was therefore only possible to train models with representation size \(200\), and DM models whose combination operation was averaging (rather than concatenation). 

\vspace{5pt}\noindent{\bf Bottom-Up Methods} We train {\bf CBOW} and {\bf SkipGram} word embeddings \cite{mikolov2013distributed} on the Books corpus, and compose by elementwise addition as proposed by \newcite{mitchell2010composition}.\footnote{We also tried multiplication but this gave very poor results.} 

We also compare to {\bf C-PHRASE}~\cite{marcobaronijointly}, an approach that exploits a (supervised) parser to infer distributed semantic representations based on a syntactic parse of sentences. C-PHRASE achieves state-of-the-art results for distributed representations on several evaluations used in this study.\footnote{Since code for C-PHRASE is not publicly-available we use the available pre-trained model ({\scriptsize \url{http://clic.cimec.unitn.it/composes/cphrase-vectors.html}}). Note this model is trained on \(3\times\) more text than others in this study.}

\vspace{5pt}\noindent{\bf Non-Distributed Baseline} We implement a {\bf TFIDF BOW} model in which the representation of sentence \(S\) encodes the count in \(S\) of a set of feature-words weighted by their \emph{tfidf} in \(C\), the corpus. The feature-words are the 200,000 most common words in \(C\). 

\subsection{Models Trained on Structured Resources}
The following models rely on (freely-available) data that has more structure than raw text.

\vspace{5pt}\noindent{\bf DictRep} \newcite{hill2015learning} trained neural language models to map dictionary definitions to pre-trained word embeddings of the words defined by those definitions. They experimented with {\bf BOW} and {\bf RNN} (with LSTM) encoding architectures and variants in which the input word embeddings were either learned or pre-trained ({\bf+embs.}) to match the target word embeddings. We implement their models using the available code and training data.\footnote{{\scriptsize \url{https://www.cl.cam.ac.uk/~fh295/}}. Definitions from the training data matching those in the WordNet STS 2014 evaluation (used in this study) were excluded.}

\vspace{5pt}\noindent{\bf CaptionRep} Using the same overall architecture, we trained ({\bf BOW} and {\bf RNN}) models to map captions in the COCO dataset~\cite{chen2015microsoft} to pre-trained vector representations of images. The image representations were encoded by a deep convolutional network \cite{szegedy2014going} trained on the ILSVRC 2014 object recognition task \cite{russakovsky2014imagenet}. Multi-modal distributed representations can be encoded by feeding test sentences forward through the trained model. 

\vspace{5pt}\noindent{\bf NMT} We consider the sentence representations learned by neural MT models. These models have identical architecture to SkipThought, but are trained on sentence-aligned translated texts. We used a standard architecture \cite{cho2014learning} on all available {\bf En-Fr} and {\bf En-De} data from the 2015 Workshop on Statistical MT (WMT).\footnote{\scriptsize \url{www.statmt.org/wmt15/translation-task.html}} 

\subsection{Novel Text-Based Models}
We introduce two new approaches designed to address certain limitations with the existing models.

\vspace{5pt}\noindent{\bf Sequential (Denoising) Autoencoders} The SkipThought objective requires training text with a coherent inter-sentence narrative, making it problematic to port to domains such as social media or artificial language generated from symbolic knowledge. To avoid this restriction, we experiment with a representation-learning objective based on \emph{denoising autoencoders} (DAEs). In a DAE, high-dimensional input data is corrupted according to some noise function, and the model is trained to recover the original data from the corrupted version. As a result of this process, DAEs learn to represent the data in terms of features that explain its important factors of variation~\cite{vincent2008extracting}. Transforming data into DAE representations (as a `pre-training' or initialisation step) gives more robust (supervised) classification performance in deep feedforward networks \cite{vincent2010stacked}.

The original DAEs were feedforward nets applied to (image) data of fixed size. Here, we adapt the approach to variable-length sentences by means of a noise function \(N(S | p_o,p_x)\), determined by free parameters \(p_o,p_x \in [0,1]\). First, for each word \(w\) in \(S\), \(N\) deletes \(w\) with (independent) probability \(p_o\). Then, for each non-overlapping bigram \(w_i w_{i+1}\) in \(S\), \(N\) swaps \(w_i\) and \(w_{i+1}\) with probability \(p_x\). We then train the same LSTM-based encoder-decoder architecture as NMT, but with the denoising objective to predict (as target) the original source sentence \(S\) given a corrupted version \(N(S |p_o,p_x)\) (as source). The trained model can then encode novel word sequences into distributed representations. We call this model the \emph{Sequential Denoising Autoencoder} ({\bf SDAE}). Note that, unlike SkipThought, SDAEs can be trained on sets of sentences in arbitrary order.   

We label the case with no noise (i.e. \(p_o = p_x = 0\) and \(N \equiv id\)) {\bf SAE}. This setting matches the method applied to text classification tasks by \newcite{dai2015semi}. The `word dropout' effect when \(p_o \geq 0\) has also been used as a regulariser for deep nets in supervised language tasks \cite{iyyer2015deep}, and for large \(p_x\) the objective is similar to word-level `debagging'~\cite{sutskever2011generating}. For the SDAE, we tuned \(p_o\), \(p_x\) on the validation set (see Section~\ref{unseval}).\footnote{We searched \(p_o,p_x \in \{0.1,0.2,0.3\}\) and observed best results with \(p_o = p_x = 0.1\).} We also tried a variant ({\bf +embs}) in which words are represented by (fixed) pre-trained embeddings. 

\vspace{5pt}\noindent{\bf FastSent} The performance of SkipThought vectors shows that rich sentence semantics can be inferred from the content of adjacent sentences. The model could be said to exploit a type of \emph{sentence-level Distributional Hypothesis}~\cite{harris1954distributional,polajnar2015exploration}. Nevertheless, like many deep neural language models, SkipThought is very slow to train (see Table~\ref{modelprops}). FastSent is a simple additive (log-linear) sentence model designed to exploit the same signal, but at much lower computational expense. Given a BOW representation of some sentence in context, the model simply predicts adjacent sentences (also represented as BOW) .

More formally, FastSent learns a source \(u_w\) and target \(v_w\) embedding for each word in the model vocabulary. For a training example  \(S_{i-1},S_i,S_{i+1}\) of consecutive sentences, \(S_i\) is represented as the sum of its source embeddings \( \mathbf{s_i} = \sum_{w \in S_i} u_w \). The cost of the example is then simply:
\begin{equation} \label{eqn1}
 \sum_{w \in S_{i-1} \cup S_{i+1}} \phi(\mathbf{s_i},v_w) 
 \end{equation}
 where \( \phi(v_1,v_2) \) is the softmax function.  

We also experiment with a variant ({\bf+AE}) in which the encoded (source) representation must predict its own words as target in addition to those of adjacent sentences. Thus in FastSent+AE, (\ref{eqn1}) becomes  \begin{equation} 
\sum_{w \in S_{i-1} \cup S_{i} \cup S_{i+1}} \phi(\mathbf{s_i},v_w).
\end{equation}

\noindent At test time the trained model (very quickly) encodes unseen word sequences into distributed representations with \( \mathbf{s} = \sum_{w \in S} u_w \).

\begin{table}

  \small
    \begin{tabular}{l|ccccccc}\\
         & \rot{OS} & \rot{R} & \rot{WO} & \rot{SD} & \rot{WD}
        & \rot{TR} & \rot{TE} \\
        \cmidrule{1-8}
        S(D)AE              & & & \OK  & 2400 & 100 & 72* &  640 \\
        ParagraphVec &  & &   & 100 & 100&  4 & 1130  \\
        CBOW                &&  &   &  500 & 500 &  2 & 145  \\
        SkipThought             & \OK &  & \OK  & 4800 & 620 & 336* & 890  \\
        FastSent               & \OK &  & & 100 & 100 & 2  & 140  \\
        DictRep                & &  \OK & \OK  & 500 & 256 & 24* & 470  \\
        CaptionRep                & & \OK  & \OK  & 500 &  256 & 24* & 470  \\
        NMT                & & \OK & \OK  &  2400 & 512 &  72* & 720  \\
        \cmidrule[1pt]{1-8}
    \end{tabular}
    \caption{\label{modelprops} {\bf Properties of models compared in this study} \hspace{6mm} {\bf OS:} requires training corpus of sentences in order. {\bf R:} requires structured resource for training. {\bf WO:} encoder sensitive to word order. {\bf SD:} dimension of sentence representation. {\bf WD:} dimension of word representation. {\bf TR:} approximate training time (hours) on the dataset in this paper. * indicates trained on GPU. {\bf TE:} approximate time (s) taken to encode 0.5m sentences.}

\end{table}
\begin{table*}[ht]
\begin{adjustwidth}{-0.6cm}{}
\renewcommand{\tabcolsep}{4.6pt}
\small
\newcommand{\mc}[1]{\multicolumn{1}{l|}{#1}}
  \begin{center}

      {
        \begin{tabular}{rr|l|l|c}
          \multicolumn{2}{c|}{Dataset} & \multicolumn{1}{c}{Sentence 1} &\multicolumn{1}{|c|}{Sentence 2} & \(/5\)  \\
          \hline
          \hline
           & News &  \emph{Mexico wishes to guarantee citizens' safety.} &\emph{Mexico wishes to avoid more violence.} & 4 \\
           & Forum &  \emph{The problem is simpler than that.} &  \emph{The problem is simple.}  & 3.8 \\
           STS & WordNet &  \emph{A social set or clique of friends.} &  \emph{An unofficial association of people or groups.} & 3.6 \\
           2014 & Twitter & \emph{Taking Aim  \#Stopgunviolence \#Congress \#NRA} & \emph{Obama, Gun Policy and the N.R.A.}  & 1.6 \\
            & Images & \emph{A woman riding a brown horse.} &  \emph{A young girl riding a brown horse.} & 4.4 \\
           & Headlines &  \emph{Iranians Vote in Presidential Election.} &  \emph{Keita Wins Mali Presidential Election.} & 0.4  \\
          \hline
          \multicolumn{2}{l|}{SICK (test+train)} & \emph{A lone biker is jumping in the air.} & \emph{A man is jumping into a full pool.}  & 1.7 \\
          \hline 
        \end{tabular}

    }
    \caption{\label{unsex} Example sentence pairs and `similarity' ratings from the unsupervised evaluations used in this study.}
  \end{center}
  \vspace*{-4ex}
  \end{adjustwidth}
\end{table*}

\subsection{Training and Model Selection}

Unless stated above, all models were trained on the Toronto Books Corpus,\footnote{\scriptsize \url{http://www.cs.toronto.edu/~mbweb/}} which has the inter-sentential coherence required for SkipThought and FastSent. The corpus consists of 70m ordered sentences from over 7,000 books. 

Specifications of the models are shown in Table~\ref{modelprops}. The log-linear models (SkipGram, CBOW, ParagraphVec and FastSent) were trained for one epoch on one CPU core. The representation dimension \(d\) for these models was found after tuning \(d \in \{100,200,300,400,500\}\) on the validation set.\footnote{For ParagraphVec only \(d \in\{100,200\}\) was possible due to the high memory footprint.} All other models were trained on one GPU. The S(D)AE models were trained for one epoch (\(\approx8\) days). The SkipThought model was trained for two weeks, covering just under one epoch.\footnote{Downloaded from {\scriptsize \url{https://github.com/ryankiros/skip-thoughts}}} For CaptionRep and DictRep, performance was monitored on held-out training data and training was stopped after 24 hours after a plateau in cost. The NMT models were trained for 72 hours. 

\section{Evaluating Sentence Representations}

In previous work, distributed representations of language were evaluated either by measuring the effect of adding representations as features in some classification task - \emph{supervised evaluation} \cite{collobert2011natural,mikolov2013efficient,kiros2015skip} - or by comparing with human relatedness judgements - \emph{unspervised evaluation} \cite{hill2015learning,baroni2014don,levy2015improving}. The former setting reflects a scenario in which representations are used to inject general knowledge (sometimes considered as \emph{pre-training}) into a supervised model. The latter pertains to applications in which the sentence representation space is used for direct comparisons, lookup or retrieval. Here, we apply and compare both evaluation paradigms.

\begin{table*}[ht]
\small
\newcommand{\mc}[1]{\multicolumn{1}{l|}{#1}}
  \begin{center}

      {
        \begin{tabular}{c|l|cccccc}
           \multicolumn{1}{c}{Data} & \multicolumn{1}{|c|}{Model} & MSRP (Acc / F1) & MR & CR & SUBJ & MPQA & TREC 
          \\
          \hline
          \hline
          & SAE & 74.3 / 81.7	& 62.6	& 68.0	& 86.1	& 76.8	& 80.2 \\
          & SAE+embs. & 70.6 / 77.9	& 73.2	& 75.3	& 89.8	& 86.2	& 80.4 \\
          Unordered  & SDAE & \bf \underline{76.4 / 83.4}	& 67.6	& 74.0	& 89.3	& 81.3	& 77.6 \\
         Sentences & SDAE+embs. & 73.7 / 80.7	& \bf 74.6	&  \bf 78.0	& \bf 90.8	& \bf 86.9	& 78.4 \\
          (Toronto Books: &ParagraphVec DBOW & 72.9	/ 81.1	& 60.2	& 66.9	& 76.3	& 70.7	& 59.4 \\
        70m sents, & ParagraphVec DM & 73.6 / 81.9	& 61.5	& 68.6	& 76.4	& 78.1 & 55.8\\
         0.9B words) &Skipgram & 69.3 / 77.2	& 73.6	& 77.3	& 89.2	& 85.0	& 82.2 \\
         &CBOW & 67.6 / 76.1	& 73.6	& 7730 & 89.1	& 85.0 & 82.2 \\
           &Unigram TFIDF & \bf 73.6 /  81.7	& 73.7	& 79.2	& 90.3	& 82.4	& \bf 85.0 \\
          \hline 
             Ordered    & SkipThought & \bf 73.0 / 82.0 & \bf 76.5 & \bf  \underline{80.1}	& \bf \underline{93.6}	& \bf 87.1	& \bf \underline{92.2} \\
       Sentences &FastSent & 72.2 / 80.3	& 70.8	& 78.4	& 88.7	&80.6 & 76.8 \\
         (Toronto Books) &FastSent+AE & 71.2 / 79.1	& 71.8	& 76.7	& 88.8	& 81.5	& 80.4  \\
          \hline 
          &NMT En to Fr & 69.1 / 77.1	& 64.7	& 70.1	& 84.9	& 81.5	& \bf 82.8 \\
        Other  &NMT En to De & 65.2 / 73.3 & 61.0 & 67.6	& 78.2	& 72.9 & 81.6 \\
          structured & CaptionRep BOW & 73.6 / 81.9	& 61.9	& 69.3	& 77.4	& 70.8	& 72.2  \\
          data & CaptionRep RNN & 72.6 / 81.1 & 55.0 & 64.9 & 64.9 & 71.0 & 62.4 \\
           resource &DictRep BOW & \bf 73.7 / 81.6	& 71.3	& 75.6	& 86.6	& 82.5 & 73.8 \\
          &DictRep BOW+embs. & 68.4 / 76.8	&  \bf \underline{76.7}	& \bf 78.7	& \bf 90.7	& \bf \underline{87.2}	& 81.0  \\
           &DictRep RNN & 73.2	/ 81.6	& 67.8	& 72.7	& 81.4	& 82.5	& 75.8 \\
           &DictRep RNN+embs. & 66.8	 / 76.0 & 72.5	& 73.5	& 85.6	& 85.7 & 72.0 \\
          \hline   

          2.8B words &CPHRASE & 72.2	 / 79.6	& 75.7	& 78.8	& 91.1	&  86.2	& 78.8 \\
           
          \hline 
        \end{tabular}

    }
    \caption{\label{supervised} Performance of sentence representation models on {\bf supervised} evaluations  (Section~\ref{supersec}). Bold numbers indicate best performance in class. Underlined indicates best overall. }
  \end{center}
  \vspace*{-4ex}
\end{table*}

\begin{table*}[ht]
\small
\newcommand{\mc}[1]{\multicolumn{1}{l|}{#1}}
  \begin{center}

      {
        \begin{tabular}{l|cccccc|c|c}
          & \multicolumn{7}{c|}{STS 2014} & SICK \\
           \multicolumn{1}{c|}{Model} & News & Forum & WordNet & Twitter & Images & Headlines & All  & \multicolumn{1}{c}{Test + Train}  \\
          \hline
\hline
    SAE & 17/.16 & .12/.12 & 	.30/.23 & 	.28/.22 & 	.49/.46 & 	.13/.11 & 	.12/.13 & .32/.31 \\
          SAE+embs. & .52/.54 & .22/.23 &  .60/.55 &  .60/.60 & . 64/.64 & .41/.41 & .42/.43 & .47/.49\\
 SDAE & .07/.04 &  .11/.13& .33/.24 & .44/.42 & .44/.38 & .36/.36 & .17/.15 & .46/.46 \\
 SDAE+embs.  & .51/.54 & .29/.29	& .56/.50	& .57/.58	& .59/.59	& .43/.44	& .37/.38 & .46/.46 \\
          ParagraphVec DBOW & .31/.34 & .32/.32 & .53/.5 & .43/.46 & .46/.44 & .39/.41 & .42/.43 & .42/.46\\
          ParagraphVec DM & .42/.46 & .33/.34 & .51/.48 & .54/.57 & .32/.30 &  .46/.47 &  .44/.44 & .44/.46 \\
                  Skipgram &.56/.59&	.42/.42& \bf	.73/.70& \underline{\bf .71}/.74&	.65/.67& {\bf	.55}/.58&	.62/.63 & \bf .60/.69 \\
          CBOW & \bf .57/.61 & \bf	.43/.44	& .72/.69	&\underline{\bf .71/.75}&.71/.73& \bf	.55/.59&	\bf .64/.65 & \bf .60/.69 \\
Unigram TFIDF & .48/.48 & .40/.38 & .60/.59  & .63/.65 & \bf 72/.74 &.49/.49	&.58/.57 & .52/.58 \\
          \hline 
           SkipThought & .44/.45 & .14/.15 & .39/.34 & .42/.43 & .55/.60 & .43/.44 & .27/.29 & .57/.60\\
FastSent &  {\bf .58/.59} & {\bf .41}/.36 & \bf .74/.70 & .63/.66 &  \bf{.74/.78} & .57/.59 &  \bf .63/.64 & \underline{\bf .61/.72}\\
FastSent+AE & .56/ \bf{.59} & \bf{.41/.40} &  .69/.64 & \bf .70/.74 & .63/.65 &  \bf{.58/.60} &  .62/.62 & .60/.65 \\
          \hline 
          NMT En to Fr & .35/.32	& .18/.18	& .47/.43 & .55/.53	& .44/.45	& .43/.43 & .43/.42 & .47/.49 \\

          NMT En to De & .47/.43 & .26/.25 & .34/.31 & .49/.45 & .44/.43 & .38/.37 & .40/.38  &.46/46 \\
          CaptionRep BOW & .26/.26 & .29/.22	& .50/.35	& .37/.31 &  \underline{ \bf .78}/.81 & .39/.36 & .46/.42 & .56/.65 \\
          CaptionRep RNN & .05/.05	& .13/.09	& .40/.33	& .36/.30	& .76/\underline{\bf .82}	& .30/.28 & .39/.36 & .53/.62\\
                    DictRep BOW & .62/.67 	&.42/.40	&.81/.81	&.62/.66	&.66/.68	&.53/.58	&.62/.65 & .57/.66\\
          DictRep BOW+embs. & \bf .65/\underline{.72}	& \bf \underline{.49/.47}	& \bf  \underline{.85/.86}	& \bf  .67/.72	&.71/.74	& \bf .57/.61	&  \bf \underline{.67/.70}  & \underline{ {\bf .61}}/.70 \\
DictRep RNN & .40/.46	&.26/.23	&.78/.78	&.42/.42	&.56/.56	&.38/.40	&.49/.50 & .49/.56 \\
DictRep RNN+embs. & .51/.60	&.29/.27	&.80/.81	&.44/.47	&.65/.70	&.42/.46	&.54/.57 & .49/.59 \\
          \hline   
          CPHRASE & \underline{\bf .69}/.71 & .43/.41 & .76/.73  & .60/.65 & .75/.79 &  \bf \underline{.60/.65} &  .65/.67 & .60/\underline{\bf .72} \\
           
          \hline 
        \end{tabular}

    }
    \caption{\label{unsupervised} Performance of sentence representation models (Spearman/Pearson correlations) on {\bf unsupervised} (relatedness) evaluations (Section~\ref{unseval}). Models are grouped according to training data as indicated in Table~\ref{supervised}.}
  \end{center}
  \vspace*{-4ex} 
\end{table*}

\subsection{Supervised Evaluations}
\label{supersec}
Representations are applied to 6 sentence classification tasks: paraphrase identification (MSRP) \cite{dolan2004unsupervised}, movie review sentiment (MR)~\cite{pang2005seeing}, product reviews (CR)~\cite{hu2004mining}, subjectivity classification (SUBJ)~\cite{pang2004sentimental}, opinion polarity (MPQA)~\cite{wiebe2005annotating} and question type classification (TREC)~\cite{voorhees2002overview}. We follow the procedure (and code) of \newcite{kiros2015skip}: a logistic regression classifier is trained on top of sentence representations, with 10-fold cross-validation used when a train-test split is not pre-defined. 

\subsection{Unsupervised Evaluations}
\label{unseval}
We also measure how well representation spaces reflect human intuitions of the semantic sentence relatedness, by computing the cosine distance between vectors for the two sentences in each test pair, and correlating these distances with gold-standard human judgements. The SICK dataset~\cite{marelli2014sick} consists of 10,000 pairs of sentences and relatedness judgements. The STS 2014 dataset~\cite{agirre2014semeval} consists of 3,750 pairs and ratings from six linguistic domains. Example ratings are shown in Table~\ref{unsex}. All available pairs are used for testing apart from the 500 SICK `trial' pairs, which are held-out for tuning hyperparameters (representation size of log-linear models, and noise parameters in SDAE). The optimal settings on this task are then applied to both supervised and unsupervised evaluations.

\section{Results}

Performance of the models on the supervised evaluations (grouped according to the data required by their objective) is shown in Table~\ref{supervised}. Overall, SkipThought vectors perform best on three of the six evaluations, the BOW DictRep model with pre-trained word embeddings performs best on two, and the SDAE on one. SDAEs perform notably well on the paraphrasing task, going beyond SkipThought by three percentage points and approaching state-of-the-art performance of models designed specifically for the task \cite{ji2013discriminative}. SDAE is also consistently better than SAE, which aligns with other findings that adding noise to AEs produces richer representations~\cite{vincent2008extracting}.  

Results on the unsupervised evaluations are shown in Table~\ref{unsupervised}. The same DictRep model performs best on four of the six STS categories (and overall) and is joint-top performer on SICK. Of the models trained on raw text, simply adding CBOW word vectors works best on STS. The best performing raw text model on SICK is FastSent, which achieves almost identical performance to C-PHRASE's state-of-the-art performance for a distributed model~\cite{marcobaronijointly}. Further, it uses less than a third of the training text and does not require access to (supervised) syntactic representations for training. Together, the results of FastSent on the unsupervised evaluations and SkipThought on the supervised benchmarks provide strong support for the sentence-level distributional hypothesis: the context in which a sentence occurs provides valuable  information about its semantics.

Across both unsupervised and supervised evaluations, the BOW DictRep with pre-trained word embeddings exhibits by some margin the most consistent performance. Ths robust performance suggests that DictRep representations may be particularly valuable when the ultimate application is non-specific or unknown, and confirms that dictionary definitions (where available) can be a powerful resource for representation learning.  

\section{Discussion}

Many additional conclusions can be drawn from the results in Tables~\ref{supervised} and ~\ref{unsupervised}. 

\vspace{5pt}\noindent{\bf Different objectives yield different representations} It may seem obvious, but the results confirm that different learning methods are preferable for different intended applications (and this variation appears greater than for word representations). For instance, it is perhaps unsurprising that SkipThought performs best on TREC because the labels in this dataset are determined by the language immediately following the represented question (i.e. the answer)~\cite{voorhees2002overview}. Paraphrase detection, on the other hand, may be better served by a model that focused entirely on the content \emph{within} a sentence, such as SDAEs. Similar variation can be observed in the unsupervised evaluations. For instance, the (multimodal) representations produced by the CaptionRep model do not perform particularly well apart from on the Image category of STS where they beat all other models, demonstrating a clear effect of the well-studied modality differences in representation learning~\cite{bruni2014multimodal}. 

The nearest neighbours in Table~\ref{neighbours} give a more concrete sense of the representation spaces. One notable difference is between (AE-style) models whose semantics come from within-sentence relationships (CBOW, SDAE, DictRep, ParagraphVec) and SkipThought/FastSent, which exploit the context around sentences. In the former case, nearby sentences generally have a high proportion of words in common, whereas for the latter it is the general concepts and/or function of the sentence that is similar, and word overlap is often minimal. Indeed, this may be a more important trait of FastSent than the marginal improvement on the SICK task. Readers can compare the CBOW and FastSent spaces at \url{http://45.55.60.98/}.

\vspace{35pt}\noindent{\bf Differences between supervised and unsupervised performance} Many of the best performing models on the supervised evaluations do not perform well in the unsupervised setting. In the SkipThought, S(D)AE and NMT models, the cost is computed based on a non-linear decoding of the internal sentence representations, so, as also observed by~\cite{almahairi2015learning}, the informative geometry of the representation space may not be reflected in a simple cosine distance. The log-linear models generally perform better in this unsupervised setting. 

\vspace{5pt}\noindent{\bf Differences in resource requirements}
As shown in Table~\ref{modelprops}, different models require different resources to train and use. This can limit their possible applications. For instance, while it was easy to make an online demo for fast querying of near neighbours in the CBOW and FastSent spaces, it was not practical for other models owing to memory footprint, encoding time and representation dimension. 
\begin{table*}[ht]
\begin{adjustwidth}{-1cm}{}
\renewcommand{\tabcolsep}{4.6pt}
\footnotesize
  \begin{center}
      {
        \begin{tabular}{l|l|l}
           \multirow{2}{*}{\bf Query} &  \bf \emph{If he had a weapon, he could maybe take out} & \bf \emph{An annoying buzz started to ring in my ears, becoming}  \\
           &\bf \emph{their last imp, and then beat up Errol and Vanessa.} & \bf  \emph{louder and louder as my vision began to swim. } \\
\hline
\hline
           \multirow{2}{*}{CBOW} &  \emph{Then Rob and I would duke it out, and every} &  \emph{Louder.}  \\
           &\emph{once in a while, he would actually beat me.} & 
\\
\hline
           \multirow{1}{*}{Skip} &  \emph{If he could ram them from behind, send them saling over } &  \emph{A weighty pressure landed on my lungs and my vision blurred} \\
           Thought&\emph{the far side of the levee, he had a chance of stopping them.} &  \emph{at the edges, threatening my consciousness altogether.}
\\
\hline
           \multirow{2}{*}{FastSent} &  \emph{Isak's close enough to pick off any one of them,} &  \emph{The noise grew louder, the quaking increased as the} \\
           &\emph{maybe all of them, if he had his rifle and a mind to. } &  \emph{sidewalk beneath my feet began to tremble even more.}
\\
\hline

           \multirow{2}{*}{SDAE} &  \emph{He'd even killed some of the most dangerous criminals} &  \emph{I smile because I'm familiar with the knock,} \\
           &\emph{in the galaxy, but none of those men had gotten to him like Vitktis.} &  \emph{pausing to take a deep breath before dashing down the stairs.}
\\
\hline
           \multirow{1}{*}{DictRep} &  \emph{Kevin put a gun to the man's head, but even though} &  \emph{Then gradually I began to hear a ringing in my ears.}  \\
         (FF+embs.)  &\emph{he cried, he couldn't tell Kevin anything more.} & \emph{}
\\\hline
           \multirow{1}{*}{Paragraph} &  \emph{I take a deep breath and open the doors.} &  \emph{They listened as the motorcycle-like roar}  \\
       Vector (DM) & & \emph{of an engine got louder and louder then stopped.}
\\
        \end{tabular}
    }
    \caption{\label{neighbours}Sample nearest neighbour queries selected from a randomly sampled 0.5m sentences of the Toronto Books Corpus.}
  \end{center}
  \vspace*{-4ex}
  \end{adjustwidth}
\end{table*}

\hspace*{-2cm}
\begin{table*}[ht]
\begin{adjustwidth}{-1cm}{}
\renewcommand{\tabcolsep}{4.6pt}
\footnotesize
\begin{center}
      {
        \begin{tabular}{cccccc|cccccccc}
          \multicolumn{6}{c|}{Supervised (combined \(\alpha = 0.90\))} & \multicolumn{8}{c}{Unsupervised (combined \(\alpha = 0.93\))} \\
          \hline
                     MSRP & MR & CR & SUBJ & MPAQ & TREC & News & Forum & WordNet & Twitter & Images & Headlines & All STS  & SICK \\
                                          \footnotesize  0.94 (6) &	0.85 (1)	 &0.86 (4)	 &0.85 (1) &	0.86	(3) &0.89 (5) &	0.92	(4) &0.92 (3)	 &0.92 (4) &	0.93 (6)  &	0.95 (8)	 &0.92 (2)  &	0.91 (1)  &0.93 (7) \\
                                          \hline
        \end{tabular}
    }
    \caption{\label{consistency} Internal consistency (Chronbach's \(\alpha\)) among evaluations when individual benchmarks are left out of the (supervised or unsupervised) cohorts. Consistency rank within cohort is in parentheses (1 = most consistent with other evaluations).}
  \end{center}
  \vspace*{-4ex}
  \end{adjustwidth}
\end{table*}

% \\   \vspace*{-4ex}\\
%\multicolumn{3}{c}{Selected from a randomly sampled 2m sentences.} \\
 % \\
%Query &  \emph{The bird flew away over the mountains and into the distance.} & \emph{What is the code for the door?}  \\
%
%\hline
%\hline
%
%           \multirow{2}{*}{FastSent} &1: \emph{\scriptsize The crow flapped it's wings and took off into the night sky, towards the northeast.} & 1: \emph{\scriptsize Is 
%           &2: \emph{ \scriptsize Birds flew up into the sky. } & 2:  \emph{ \scriptsize Can you see the lock from through the Louvers?}
%
%\\
%
%\hline
%           \multirow{2}{*}{CBOW} &1:  \emph{ \scriptsize Helena looked away, into the distance. } & 1: \emph{ \scriptsize The code? }  \\
%           &2: \emph{ \scriptsize In the distance, the wilderness stretched away from the city.}  &  2: \emph{ \scriptsize The door?}
%
%\\
%\hline
%           \multirow{2}{*}{DictRep (FF+embs.)} &1:  \emph{ \scriptsize The beasts flew over the mountain tops like flies circling a forgotten plate of food.} & 1: \emph{ \scriptsize "What's in the door?" }  \\
%           &2: \emph{ \scriptsize Look at that big bird flying up there!}  &  2: \emph{ \scriptsize "Someone is at the door!"}
%
%\\
%\hline
%          \hline
%        \end{tabular}
%    }
%    \caption{Sample nearest neighbour queries selected from the Toronto Books Corpus.}
%  \end{center}
%  \vspace*{-4ex}
%\end{table*}

\vspace{5pt}\noindent{\bf The role of word order is unclear} 
The average scores of models that are sensitive to word order (76.3) and of those that are not (76.6) are approximately the same across supervised evaluations. Across the unsupervised evaluations, however, BOW models score 0.55 on average compared with 0.42 for RNN-based (order sensitive) models. This seems at odds with the widely held view that word order plays an important role in determining the meaning of English sentences. One possibility is that order-critical sentences that cannot be disambiguated by a robust conceptual semantics (that could be encoded in distributed lexical representations) are in fact relatively rare. However, it is also plausible that current available evaluations do not adequately reflect order-dependent aspects of meaning (see below). This latter conjecture is supported by the comparatively strong performance of TFIDF BOW vectors, in which the effective lexical semantics are limited to simple relative frequencies.  

\vspace{5pt}\noindent{\bf The evaluations have limitations} The internal consistency (Chronbach's \(\alpha\)) of all evaluations considered together is \(0.81\) (just above `acceptable').\footnote{\scriptsize \url{wikipedia.org/wiki/Cronbach's_alpha}} Table~\ref{consistency} shows that consistency is far higher (`excellent') when considering the supervised or unsupervised tasks as independent cohorts. This indicates that, with respect to common characteristics of sentence representations, the supervised and unsupervised benchmarks do indeed prioritise different properties. It is also interesting that, by this metric, the properties measured by MSRP and image-caption relatedness are the furthest removed from other evaluations in their respective cohorts.

While these consistency scores are a promising sign, they could also be symptomatic of a set of evaluations that are all limited in the same way. The inter-rater agreement is only reported for one of the 8 evaluations considered (MPQA, \(0.72\)~\cite{wiebe2005annotating}), and for MR, SUBJ and TREC, each item is only rated by one or two annotators to maximise coverage. Table~\ref{unsex} illustrates why this may be an issue for the unsupervised evaluations; the notion of sentential 'relatedness' seems very subjective. It should be emphasised, however, that the tasks considered in this study are all frequently used for evaluation, and, to our knowledge, there are no existing benchmarks that overcome these limitations. 

%The evaluation of representation-learning algorithms is notoriously problematic. It remains to be determined to what extent the %relatively strong performance of simple orderless and even non-distributed representations is caused by limitations in the %evaluations versus weaknesses in the models. In any case, a key challenge for future research is the development of benchmarks %that reflect robust, consistent semantic or cognitive effects, particularly above the word level.

%\vspace{5pt}\noindent{\bf Sentence representations may not be desirable} While most factual or proposition information in language seems to reside in multi-word chunks, sentences may not be a cognitively realistic or practically useful scope for distributed representations. Indeed various recent studies with neural language models report optimal performance with models that can learn to focus on phrase-like (multiple-word) sub-chunks of sentences rather than those considering entire sentences `at once'~\cite{hill2015goldilocks,luong2015effective}. Both the difficulty mining consistent human opinions and the fact that extremely unexpressive models (Unigram TFIDF) perform competitively at the sentence level is consistent with this conclusion. Of course, studying representations at a level between the word sentence level poses the challenge of how to delimit such chunks in a relatively objective or unbiased way. We leave this conundrum for future work. 

\section{Conclusion}
Advances in deep learning algorithms, software and hardware mean that many architectures and objectives for learning distributed sentence representations from unlabelled data are now available to NLP researchers. We have presented the first (to our knowledge) systematic comparison of these methods. We showed notable variation in the performance of approaches across a range of evaluations. Among other conclusions, we found that the optimal approach depends critically on whether representations will be applied in supervised or unsupervised settings - in the latter case, fast, shallow BOW models can still achieve the best performance. Further, we proposed two new objectives, FastSent and Sequential Denoising Autoencoders, which perform particularly well on specific tasks (MSRP and SICK sentence relatedness respectively).\footnote{We make all code for training and evaluating these new models publicly available, together with pre-trained models and an online demo of the FastSent sentence space.} If the application is unknown, however, the best all round choice may be DictRep: learning a mapping of pre-trained word embeddings from the word-phrase signal in dictionary definitions. While we have focused on models using naturally-occurring training data, in future work we will also consider supervised architectures (including convolutional, recursive and character-level models), potentially training them on multiple supervised tasks as an alternative way to induce the 'general knowledge' needed to give language technology the elusive human touch. 

\section*{Acknowledgments}

This work was supported by a Google Faculty Award to AK and FH and a Google European Doctoral Fellowship to FH. Thanks also to Marek Rei, Tamara Polajnar, Laural Rimell, Jamie Ryan Kiros and Piotr Bojanowski for helpful discussion and comments. 

\bibliography{naaclhlt2016}
\bibliographystyle{naaclhlt2016}
\end{document}